\pdfoutput=1
\documentclass[11pt]{article}

\usepackage[]{naacl2021}

\usepackage{graphicx}

\usepackage{times}
\usepackage{latexsym}

\usepackage[T1]{fontenc}
\usepackage[utf8]{inputenc}

\usepackage{microtype}

\usepackage{multirow}
\usepackage{multicol}
\usepackage{booktabs}
\usepackage{float}
\usepackage{subfigure}

\newcommand*{\affaddr}[1]{#1} % No op here. Customize it for different styles.
\newcommand*{\affmark}[1][*]{\textsuperscript{#1}}
\newcommand*{\email}[1]{\texttt{#1}}

\RequirePackage{color}

\title{Stress Test Evaluation of Biomedical Word Embeddings}% for Biomedical Named Entity Recognition}
\author{%
Vladimir Araujo\affmark[1,2], Andrés Carvallo\affmark[1,2], Carlos Aspillaga\affmark[1], Camilo Thorne\affmark[3], Denis Parra\affmark[1,2]\\
\affaddr{\affmark[1]Pontificia Universidad Católica de Chile}\\
\affaddr{\affmark[2]Millennium Institute for Foundational Research on Data
(IMFD)}\\
\affaddr{\affmark[3]Elsevier}\\
\email{\{vgaraujo,afcarvallo,cjaspill\}@uc.cl}\\
\email{c.thorne.1@elsevier.com}\\
\email{dparra@ing.puc.cl}
}
     
\begin{document}
\maketitle
\begin{abstract}
 The success of pretrained word embeddings has motivated their use in the biomedical domain, with contextualized embeddings yielding remarkable results in several biomedical NLP tasks. However, there is a lack of research on quantifying their behavior under severe ``stress'' scenarios. In this work, we systematically evaluate three language models with adversarial examples -- automatically constructed tests that allow us to examine how robust the models are. We propose two types of stress scenarios focused on the biomedical named entity recognition (NER) task, one inspired by spelling errors and another based on the use of synonyms for medical terms. Our experiments with three benchmarks show that the performance of the original models decreases considerably, in addition to revealing their weaknesses and strengths. Finally, we show that adversarial training causes the models to improve their robustness and even to exceed the original performance in some cases.
\end{abstract}

%%%%%%%%%%%%%%%%%%%%%%%%%%%%%%%%%%%%%%%%
\section{Introduction}% and Related Work}
%%%%%%%%%%%%%%%%%%%%%%%%%%%%%%%%%%%%%%%%

Biomedical NLP (BioNLP) is the field concerned with developing NLP tools and methods for the life sciences domain. Some applications of these techniques include e.g., discovery of gene-disease interactions \cite{pletscher2015diseases}, development of new drugs \cite{tari2010discovering}, or automatic screening of biomedical documents  \cite{Carvallo2020}. With the exponential growth of digital biomedical literature, the importance of BioNLP has become especially relevant as a tool to extract relevant knowledge for making decisions in clinical settings as well as in public health. 
In order to encourage the development of this area, public datasets and challenges have been shared with the community to solve these tasks, such as BioSSES \cite{souganciouglu2017biosses}, HOC \cite{hoc2000}, ChemProt \cite{chemprot2016} and BC5CDR \cite{Li2016}, among others. At the same time, neural language models have shown significant progress since the introduction of models such as W2V \cite{NIPS2013_5021}, and more recent models like ELMo \cite{peters2018deep} and BERT \cite{devlin-etal-2019-bert}. These models, trained over large corpora (MEDLINE and PubMed in the biomedical domain) have obtained remarkable results in most NLP tasks, including BioNLP benchmarks \cite{peng-etal-2019-transfer}. 
However, they have not been systematically evaluated under severe stress conditions to test their robustness to specific linguistic phenomena.
For this reason, the objective of this paper is to evaluate three well-known neural language models under stress conditions. As a case study, we evaluate NER benchmarks since it a key BioNLP information extraction task.
% Concretely, we carry out an evaluation of well-known language models on a number of adversarially perturbed disease and chemical NER benchmarks. 

Our stress test evaluation is inspired by the work of \citet{naik-etal-2018-stress}, which proposes the use of adversarial evaluation for natural language inference by adding distractions in sentences, and evaluating models on this test set.
We propose an adversarial evaluation black-box methodology, which does not require access to the inner workings of the models in order to generate adversarial examples \cite{zhang2019adversarial}. Specifically, we make perturbations to the input data, also known as edit adversaries, that could cause the models to fall into erroneous predictions. 
% This methodology is inspired by the work of \citet{naik-etal-2018-stress}, which proposes the use of adversarial evaluation for NLI by adding distractions in sentences, and evaluating models on this test set.
Additionally, we train the models with the proposed adversarial examples, which is a methodology used in previous works \cite{belinkov2018synthetic,jia-liang-2017-adversarial} to strengthen the neural language models during the training process. We hope that our work will motivate the development and use of adversarial examples to evaluate models and obtain more robust biomedical embeddings.

%%%%%%%%%%%%%%%%%%%%%%%%%%%%%%%%%%%%%%%%
%%%%%%%%%%%%%%%%%%%%%%%%%%%%%%%%%%%%%%%%
% Table 1
\begin{table*}[!tb]
% \small
\addtolength{\tabcolsep}{-1pt}
\renewcommand{\arraystretch}{1.1}
\fontsize{9.8pt}{11pt}\selectfont
\centering
\begin{tabular}{@{}|l|l|@{}}
% \toprule
\hline
\textbf{Original (O)} & Linoleic acid autoxidation inhibitions on all fractions were higher than that on alpha-tocopherol. \\ \hline
\textbf{Keyboard (K)} & 
Linoleic \textcolor{purple}{avid autoxidatiob inh9bitions} on all \textcolor{purple}{fractjons} were higher than that on \textcolor{purple}{zlpha-toclpherol}. \\ \hline
\textbf{Swap (W)} & Linoleic \textcolor{purple}{aicd autoxidtaion inhibtiions} on all \textcolor{purple}{fractoins} were higher than that on \textcolor{purple}{aplha-tocohperol}.\\ \hline
\textbf{Synonymy (S)} & Linoleic acid autoxidation inhibitions on all fractions were higher than that on \textcolor{purple}{vitamin E}. \\  \hline
% \bottomrule
\end{tabular}
\vspace{-3mm}
\caption{\label{table-1} Examples of sentences of the stress tests.}
\vspace{-4mm}
\end{table*}
%%%%%%%%%%%%%%%%%%%%%%%%%%%%%%%%%%%%%%%%
%%%%%%%%%%%%%%%%%%%%%%%%%%%%%%%%%%%%%%%%

%%%%%%%%%%%%%%%%%%%%%%%%%%%%%%%%%%%%%%%%
\section{Related Work}
%%%%%%%%%%%%%%%%%%%%%%%%%%%%%%%%%%%%%%%%

%%%%%%%%%%%%%%%%%%%%%%%%%%%%%%%%%%%%%%%%
\paragraph{Adversarial Evaluation of NLP Models}
%%%%%%%%%%%%%%%%%%%%%%%%%%%%%%%%%%%%%%%%
One way to test NLP models is by using adversarial tests, which consist of applying intentional disturbances to a gold standard, to test whether the attack leads the models into incorrect predictions. Previous works on adversarial attacks have demonstrated how dangerous it can be to use machine learning systems in real-world applications \cite{42503,goodfellow2014explaining}. Indeed, it is known that even small amounts of noise can cause severe failures in neural computer vision models \cite{Akhtar_2018}. However, such failures can be mitigated through adversarial training \cite{goodfellow2014explaining}. These properties have in turn motivated novel adversarial strategies designed for various NLP tasks \cite{zhang2019adversarial}, as well as work on adversarial attacks focused on recurrent and transformer networks applied to \textit{generic} NLP benchmarks \cite{aspillaga-etal-2020-stress}. 

%%%%%%%%%%%%%%%%%%%%%%%%%%%%%%%%%%%%%%%%
\paragraph{Evaluation of Biomedical Models} 
%%%%%%%%%%%%%%%%%%%%%%%%%%%%%%%%%%%%%%%%
Models used in BioNLP tasks elicit particular interest in this context because an erroneous prediction can potentially be very harmful in practice -- e.g., put at risk the health of patients \cite{10.1145/3219819.3219909}. Although adversarial attacks have been widely studied in tasks related to image analysis \cite{Paschali2018,Finlayson2019,ma2019understanding}, to the best of our knowledge, a gap still exists regarding BioNLP models and tasks \cite{winlp}.

%%%%%%%%%%%%%%%%%%%%%%%%%%%%%%%%%%%%%%%%
% \input{table1}
%%%%%%%%%%%%%%%%%%%%%%%%%%%%%%%%%%%%%%%%
% Table 2
\begin{table*}[tb]
\small
\renewcommand{\arraystretch}{1.1}
\centering
\begin{tabular}{@{}|l|l|l|l|l|l|l|@{}}
\hline
\multicolumn{1}{|c|}{\textbf{Train / Test}} & \multicolumn{1}{c|}{\textbf{Entity}} & \multicolumn{1}{c|}{\textbf{\# of sentences (annotated)}} & \multicolumn{1}{c|}{\textbf{\# of tokens}} & \multicolumn{1}{c|}{\textbf{\% K}} & \multicolumn{1}{c|}{\textbf{\% W}} & \multicolumn{1}{c|}{\textbf{\% S}} \\ \hline
BC5CDR                                         & Chemical                             & 4560 (1609) / 4797 (1706)                                 & 122730 /129547                             & 36.3 / 36.1                        & 33.7 / 33.2                        & 6.8 / 6.5                          \\ \hline
BC5CDR                                         & Disease                              & 4560 (1902) / 4797 (1955)                                 & 122730 /129547                             & 36.3 / 36.1                        & 33.7 / 33.2                        & 10.6 / 9.9                         \\ \hline
BC4CHEMD                                       & Chemical                             & 30681 (16175) / 26363 (13935)                             & 922609 / 792369                            & 37.8 / 37.6                        & 33.9 / 33.9                        & 5.2 / 5.3                          \\ \hline
NCBI-Disease                                   & Disease                              & 5423 (2501) / 939 (401)                                   & 141092 /  25397                            & 37.4 / 37.5                        & 33.4 / 33.3                        & 9.2 / 8.6                          \\ \hline
\end{tabular}
\vspace{-3mm}
\caption{\label{table-2} Details of the datasets used. The last three columns present the percentage of tokens modified for each of the adversarial datasets. The slash separates the values belonging to the training and the test set.}
\vspace{-1mm}
\end{table*}
%%%%%%%%%%%%%%%%%%%%%%%%%%%%%%%%%%%%%%%%
%%%%%%%%%%%%%%%%%%%%%%%%%%%%%%%%%%%%%%%%

%%%%%%%%%%%%%%%%%%%%%%%%%%%%%%%%%%%%%%%%
\section{Methodology}
%%%%%%%%%%%%%%%%%%%%%%%%%%%%%%%%%%%%%%%%
We follow a black-box attack methodology \cite{zhang2019adversarial}, which consists of making alterations in the input data to cause erroneous predictions in the models. 
The following subsections describe each of the adversarial sets, and their construction\footnote{All stress tests available at  \href{https://github.com/ialab-puc/BioNLP-StressTest}{https://github.com/ialab-puc/BioNLP-StressTest}.}.
We show examples of the stress tests in Table~\ref{table-1}.
%%%%%%%%%%%%%%%%%%%%%%%%%%%%%%%%%%%%%%%%
\paragraph{Noise Adversaries} 
%%%%%%%%%%%%%%%%%%%%%%%%%%%%%%%%%%%%%%%%
These adversaries test the robustness of models to \emph{spelling errors}. Inspired by \cite{belinkov2018synthetic}, we constructed adversarial examples that try to emulate spelling errors made by human beings. We used SpaCy models \cite{Neumann2019ScispaCyFA} to retrieve the medical words of each corpus and add noise to them.
We used two types of alterations: i)~\textbf{Keyboard typo noise (K)} involves replacing a random character in each relevant word with an adjacent character on QWERTY English keyboards. This methodology could be adapted to keyboards with other designs or languages. ii)~\textbf{Swap noise (W)} consists of selecting a random pair of consecutive characters in each relevant word and then swapping them. 
% See Table~\ref{table-1} for examples.

% These examples test the robustness of models to \emph{spelling errors}. Inspired by \citet{belinkov2018synthetic}, we constructed adversarial examples that try to emulate spelling errors committed by human beings\footnote{All stress tests available in  \href{https://drive.google.com/drive/folders/1M5Fdkc1VcmBnRVPKO2Y9vSQkM7aUQCXy}{our repository}.}. We used SpaCy models \cite{Neumann2019ScispaCyFA} to retrieve the medical content words of each corpus. Then, the noise was added to each entity term. We used two types of alterations: (i)~\textbf{Keyboard typo noise (K)} involves replacing a random character in each relevant word with an adjacent character on traditional English keyboard. (ii)~\textbf{Swap noise (W)} consists of selecting a random pair of consecutive characters in each relevant word and then swapping them. See Table~\ref{table-1} for examples.

%%%%%%%%%%%%%%%%%%%%%%%%%%%%%%%%%%%%%%%%
\paragraph{Synonymy Adversaries (S)} 
%%%%%%%%%%%%%%%%%%%%%%%%%%%%%%%%%%%%%%%%
These adversaries test if a model can \emph{understand synonymy relations}. 
Unlike the noise adversaries, this set focuses on modifying chemical and disease words (entities).
% For simplicity, we focused only on chemical and disease words (entities). 
We used PyMedTermino \cite{Lamy}, which uses the vocabulary of UMLS \cite{Bodenreider2004}, to find the most similar or related term (synonym) to 
% the original word. 
a certain word.
% Finally, we replaced the synonym found if it is a disease or chemical. 
If a synonym is retrieved, the original word is replaced; otherwise, it remains the same.
% See Table~\ref{table-1} for examples.
In some cases, this method changes a simple entity (one word) to a composite one (multiple words), so the gold labels are also adjusted to avoid a mismatch in the dataset.

%%%%%%%%%%%%%%%%%%%%%%%%%%%%%%%%%%%%%%%%
\paragraph{Task and Datasets} 
%%%%%%%%%%%%%%%%%%%%%%%%%%%%%%%%%%%%%%%%
Biomedical NER is the task that aims at detecting biomedical entities of interest such as proteins, cell types, chemicals, or diseases in biomedical documents. We conducted our evaluation on three biomedical NER benchmarks using the IOB2 tag format \cite{ramshaw1999text}. The \textbf{BC5CDR} corpus \cite{Li2016} is composed of mentions of chemicals and diseases found in 1,500 PubMed articles. The \textbf{BC4CHEMD} corpus \cite{Krallinger2015} contains mentions of chemicals and drugs from 10,000 MEDLINE abstracts. The \textbf{NCBI-Disease} corpus \cite{Doan2014} consists of 793 PubMed abstracts annotated with disease mentions. Table~\ref{table-2} lists the datasets used in this work along with their most relevant statistics.

%%%%%%%%%%%%%%%%%%%%%%%%%%%%%%%%%%%%%%%%
%%%%%%%%%%%%%%%%%%%%%%%%%%%%%%%%%%%%%%%%
% Table 3
\begin{table*}[tb]
\setlength{\tabcolsep}{3.3pt}
\renewcommand{\arraystretch}{1.05}

\definecolor{mygray}{rgb}{0.23, 0.27, 0.29}

\centering
\small
\begin{tabular}{@{}|@{~}l|c|c|c|c|c|c|c|c|c|c|c|c|c|c|c|c|@{}}
\hline
\multicolumn{1}{|l|}{\multirow{2}{1pt}{\textbf{Model}}} & \multicolumn{4}{c|}{\textbf{BC5CDR-Chemical}}     & \multicolumn{4}{c|}{\textbf{BC5CDR-Disease}}      & \multicolumn{4}{c|}{\textbf{BC4CHEMD}}            & \multicolumn{4}{c|}{\textbf{NCBI-Disease}}        \\ 
\cline{2-17} 
\multicolumn{1}{|c|}{}                                & \textbf{O} & \textbf{K} & \textbf{W} & \textbf{S} & \textbf{O} & \textbf{K} & \textbf{W} & \textbf{S} & \textbf{O} & \textbf{K} & \textbf{W} & \textbf{S} & \textbf{O} & \textbf{K} & \textbf{W} & \textbf{S} \\ 
\hline
\multirow{2}{1pt}{BioBERT}                                               & .937       & .745      & .635      & .770      & .863      & .407      & .473      & .366      &  .919          &   .585         &  .675          &     .678       &    .887        &  .483          &  .628          &     .683       \\ 
 &
\color{mygray}\scriptsize{${\pm}$.004} &
\color{mygray}\scriptsize{${\pm}$.006} &
\color{mygray}\scriptsize{${\pm}$.008} &
\color{mygray}\scriptsize{${\pm}$.011} &
\color{mygray}\scriptsize{${\pm}$.004} &
\color{mygray}\scriptsize{${\pm}$.008} &
\color{mygray}\scriptsize{${\pm}$.010} &
\color{mygray}\scriptsize{${\pm}$.007} &
\color{mygray}\scriptsize{${\pm}$.004} &
\color{mygray}\scriptsize{${\pm}$.005} &
\color{mygray}\scriptsize{${\pm}$.007} &
\color{mygray}\scriptsize{${\pm}$.009} &
\color{mygray}\scriptsize{${\pm}$.004} &
\color{mygray}\scriptsize{${\pm}$.007} &
\color{mygray}\scriptsize{${\pm}$.011} &
\color{mygray}\scriptsize{${\pm}$.006} \\ 
\hline
\multirow{2}{1pt}{BlueBERT}                                              & .901       & .583      & .708      & .739      & .838      & .368      & .441      & .362      &    .820        & .472           & .570           &     .607       &    .773        &  .332          &      .438      &     .615       \\ 
 &
\color{mygray}\scriptsize{${\pm}$.003} &
\color{mygray}\scriptsize{${\pm}$.005} &
\color{mygray}\scriptsize{${\pm}$.008} &
\color{mygray}\scriptsize{${\pm}$.010} &
\color{mygray}\scriptsize{${\pm}$.004} &
\color{mygray}\scriptsize{${\pm}$.007} &
\color{mygray}\scriptsize{${\pm}$.011} &
\color{mygray}\scriptsize{${\pm}$.007} &
\color{mygray}\scriptsize{${\pm}$.003} &
\color{mygray}\scriptsize{${\pm}$.004} &
\color{mygray}\scriptsize{${\pm}$.009} &
\color{mygray}\scriptsize{${\pm}$.010} &
\color{mygray}\scriptsize{${\pm}$.003} &
\color{mygray}\scriptsize{${\pm}$.006} &
\color{mygray}\scriptsize{${\pm}$.009} &
\color{mygray}\scriptsize{${\pm}$.006} \\ 
\hline

\multirow{2}{1pt}{BERT} & .887 & .563 & .684 & .738 & .816 & .356 & .431 & .336 & .808 & .443 & .509 & .598 & .771 & .305 & .433 & .583 \\
&
\color{mygray}\scriptsize{${\pm}$.004} &
\color{mygray}\scriptsize{${\pm}$.007} &
\color{mygray}\scriptsize{${\pm}$.010} &
\color{mygray}\scriptsize{${\pm}$.015} &
\color{mygray}\scriptsize{${\pm}$.006} &
\color{mygray}\scriptsize{${\pm}$.009} &
\color{mygray}\scriptsize{${\pm}$.013} &
\color{mygray}\scriptsize{${\pm}$.008} &
\color{mygray}\scriptsize{${\pm}$.004} &
\color{mygray}\scriptsize{${\pm}$.006} &
\color{mygray}\scriptsize{${\pm}$.008} &
\color{mygray}\scriptsize{${\pm}$.013} &
\color{mygray}\scriptsize{${\pm}$.005} &
\color{mygray}\scriptsize{${\pm}$.008} &
\color{mygray}\scriptsize{${\pm}$.014} &
\color{mygray}\scriptsize{${\pm}$.007} \\
\hline
\multirow{2}{1pt}{BioELMo} & .923 & .838 & .726 & .757 & .845 & .656 & .482 & .408 & .915 & .770 & .634 & .668 & .869 & .711 & .543 & .677 \\
&
\color{mygray}\scriptsize{${\pm}$.001} &
\color{mygray}\scriptsize{${\pm}$.003} &
\color{mygray}\scriptsize{${\pm}$.010} &
\color{mygray}\scriptsize{${\pm}$.032} &
\color{mygray}\scriptsize{${\pm}$.002} &
\color{mygray}\scriptsize{${\pm}$.018} &
\color{mygray}\scriptsize{${\pm}$.025} &
\color{mygray}\scriptsize{${\pm}$.013} &
\color{mygray}\scriptsize{${\pm}$.001} &
\color{mygray}\scriptsize{${\pm}$.003} &
\color{mygray}\scriptsize{${\pm}$.004} &
\color{mygray}\scriptsize{${\pm}$.004} &
\color{mygray}\scriptsize{${\pm}$.005} &
\color{mygray}\scriptsize{${\pm}$.017} &
\color{mygray}\scriptsize{${\pm}$.026} &
\color{mygray}\scriptsize{${\pm}$.012} \\ 
\hline
\multirow{2}{1pt}{ChemPatent ELMo} & .910 & .822 & .745 & .757 & .824 & .637 & .508 & .380 & .898 & .766 & .662 & .642 & .863 & .693 & .586 & .655 \\
&
\color{mygray}\scriptsize{${\pm}$.001} &
\color{mygray}\scriptsize{${\pm}$.004} &
\color{mygray}\scriptsize{${\pm}$.005} &
\color{mygray}\scriptsize{${\pm}$.016} &
\color{mygray}\scriptsize{${\pm}$.001} &
\color{mygray}\scriptsize{${\pm}$.013} &
\color{mygray}\scriptsize{${\pm}$.013} &
\color{mygray}\scriptsize{${\pm}$.017} &
\color{mygray}\scriptsize{${\pm}$.001} &
\color{mygray}\scriptsize{${\pm}$.003} &
\color{mygray}\scriptsize{${\pm}$.005} &
\color{mygray}\scriptsize{${\pm}$.005} &
\color{mygray}\scriptsize{${\pm}$.004} &
\color{mygray}\scriptsize{${\pm}$.018} &
\color{mygray}\scriptsize{${\pm}$.020} &
\color{mygray}\scriptsize{${\pm}$.009} \\ 
\hline

\multirow{2}{1pt}{ELMo} & .879 & .702 & .637 & .720 & .800 & .461 & .373 & .378 & .866 & .612 & .507 & .611 & .848 & .575 & .495 & .643 \\
&
\color{mygray}\scriptsize{${\pm}$.002} &
\color{mygray}\scriptsize{${\pm}$.010} &
\color{mygray}\scriptsize{${\pm}$.017} &
\color{mygray}\scriptsize{${\pm}$.018} &
\color{mygray}\scriptsize{${\pm}$.003} &
\color{mygray}\scriptsize{${\pm}$.023} &
\color{mygray}\scriptsize{${\pm}$.020} &
\color{mygray}\scriptsize{${\pm}$.014} &
\color{mygray}\scriptsize{${\pm}$.001} &
\color{mygray}\scriptsize{${\pm}$.007} &
\color{mygray}\scriptsize{${\pm}$.011} &
\color{mygray}\scriptsize{${\pm}$.005} &
\color{mygray}\scriptsize{${\pm}$.004} &
\color{mygray}\scriptsize{${\pm}$.034} &
\color{mygray}\scriptsize{${\pm}$.023} &
\color{mygray}\scriptsize{${\pm}$.008} \\ 
\hline
\multirow{2}{1pt}{BioMedical W2V}                                        & .873       & .231       & .238       & .719       & .788       & .132       & .133       & .351       & .846       & .233       & .244       & .589       & .827       & .284       & .292       & .596       \\ 
 &
\color{mygray}\scriptsize{${\pm}$.004} &
\color{mygray}\scriptsize{${\pm}$.012} &
\color{mygray}\scriptsize{${\pm}$.021} &
\color{mygray}\scriptsize{${\pm}$.016} &
\color{mygray}\scriptsize{${\pm}$.008} &
\color{mygray}\scriptsize{${\pm}$.009} &
\color{mygray}\scriptsize{${\pm}$.011} &
\color{mygray}\scriptsize{${\pm}$.015} &
\color{mygray}\scriptsize{${\pm}$.005} &
\color{mygray}\scriptsize{${\pm}$.008} &
\color{mygray}\scriptsize{${\pm}$.013} &
\color{mygray}\scriptsize{${\pm}$.012} &
\color{mygray}\scriptsize{${\pm}$.005} &
\color{mygray}\scriptsize{${\pm}$.014} &
\color{mygray}\scriptsize{${\pm}$.019} &
\color{mygray}\scriptsize{${\pm}$.021} \\ 
\hline
\multirow{2}{1pt}{ChemPatent W2V}                                        & .871       & .224       & .221       & .715       & .772       & .127       & .122       & .347       & .828       & .253       & .260       & .584       & .816       & .269       & .252       & .582       \\ 
{} {} {} {} {} {} {} {} {} {} {} {} {} {} {} {} {} {} {} {} {} {} {} &
\color{mygray}\scriptsize{${\pm}$.003} &
\color{mygray}\scriptsize{${\pm}$.011} &
\color{mygray}\scriptsize{${\pm}$.012} &
\color{mygray}\scriptsize{${\pm}$.015} &
\color{mygray}\scriptsize{${\pm}$.007} &
\color{mygray}\scriptsize{${\pm}$.005} &
\color{mygray}\scriptsize{${\pm}$.009} &
\color{mygray}\scriptsize{${\pm}$.016} &
\color{mygray}\scriptsize{${\pm}$.007} &
\color{mygray}\scriptsize{${\pm}$.009} &
\color{mygray}\scriptsize{${\pm}$.010} &
\color{mygray}\scriptsize{${\pm}$.012} &
\color{mygray}\scriptsize{${\pm}$.007} &
\color{mygray}\scriptsize{${\pm}$.021} &
\color{mygray}\scriptsize{${\pm}$.019} &
\color{mygray}\scriptsize{${\pm}$.013} \\ 
\hline

\multirow{2}{1pt}{W2V} & .818 & .237 & .227 & .641 & .760 & .120 & .120 & .341 & .766 & .264 & .260 & .513 & .785 & .281 & .271 & .526 \\
&
\color{mygray}\scriptsize{${\pm}$.004} &
\color{mygray}\scriptsize{${\pm}$.013} &
\color{mygray}\scriptsize{${\pm}$.013} &
\color{mygray}\scriptsize{${\pm}$.017} &
\color{mygray}\scriptsize{${\pm}$.003} &
\color{mygray}\scriptsize{${\pm}$.008} &
\color{mygray}\scriptsize{${\pm}$.009} &
\color{mygray}\scriptsize{${\pm}$.013} &
\color{mygray}\scriptsize{${\pm}$.007} &
\color{mygray}\scriptsize{${\pm}$.011} &
\color{mygray}\scriptsize{${\pm}$.012} &
\color{mygray}\scriptsize{${\pm}$.008} &
\color{mygray}\scriptsize{${\pm}$.005} &
\color{mygray}\scriptsize{${\pm}$.022} &
\color{mygray}\scriptsize{${\pm}$.019} &
\color{mygray}\scriptsize{${\pm}$.009} \\ 
\hline
\end{tabular}
\vspace{-3mm}
\caption{\label{table-3} Stress test evaluation results in terms of terms F1-score for each model and dataset. We report means and standard deviations by training and evaluating ten times with different seeds.}
\vspace{-4mm}
\end{table*}
%%%%%%%%%%%%%%%%%%%%%%%%%%%%%%%%%%%%%%%%
%%%%%%%%%%%%%%%%%%%%%%%%%%%%%%%%%%%%%%%%

%%%%%%%%%%%%%%%%%%%%%%%%%%%%%%%%%%%%%%%%
\paragraph{Embeddings and NER Models} 
%%%%%%%%%%%%%%%%%%%%%%%%%%%%%%%%%%%%%%%%
We evaluated both word (W2V) and contextualized embeddings. 
On the one hand, we assessed BioMedical W2V \cite{Pyysalo:2013b} and ChemPatent W2V \cite{zhai2019improving}. 
The ChemPatent embeddings were trained on a 1.1 billion word corpus of chemical patents from 7 patent offices, whereas all the other embeddings were trained on the PubMed corpus.
On the other hand, we evaluated BioBERT v1.1 \cite{10.1093/bioinformatics/btz682} and BlueBERT (P) \cite{peng-etal-2019-transfer}, both in their base version for convenience. BioBERT embeddings were trained on PubMed abstracts and full-text corpora consisting of 4.3 billion and 13.5 billion words each. BlueBERT was trained on 4 billion words from PubMed abstracts.
We used the implementation provided by \citet{peng-etal-2019-transfer} for NER  with default hyperparameters.\footnote{\href{https://github.com/ncbi-nlp/bluebert}{https://github.com/ncbi-nlp/bluebert}}
Finally, we evaluate BioELMo \cite{jin2019probing} and ChemPatent ELMo \cite{zhai2019improving}.
As NER models we either (a) fine-tuned BERT as proposed by \citet{peng-etal-2019-transfer} or
(b) used AllenNLP's basic biLSTM-CRF implementation\footnote{\href{https://github.com/allenai/allennlp-models/tree/main/training_config/tagging}{https://github.com/allenai/allennlp-models}}, with no hyperparameter tuning other than changing the initial embedding layer with one of the ELMo or W2V embeddings. 
For comparison purposes, we also include the ``vanilla'' version of the models mentioned above, which are pretrained with general corpora.
We trained each model 10 times using different random seeds, for 15 epochs every time. We use CoNLL evaluation \cite{agirre2007semeval}, reporting the F1 score for all datasets.

% The task consists basically of classifying document words into labels B, O, or I, which correspond to the beginning of chunk, out of the chunk, or inside the chunk, respectively, where a chunk is a set of words that refers to chemical or disease depending on the dataset.

%\camilo{Hasta aqui todo bien. La seccion 3 la veo despues.}

%%%%%%%%%%%%%%%%%%%%%%%%%%%%%%%%%%%%%%%%
\section{Experiments}

In this section we report the results of our experiments. Note that all percentage drops or increases are expressed relative to the original score, not as percentage points.
%%%%%%%%%%%%%%%%%%%%%%%%%%%%%%%%%%%%%%%%

%%%%%%%%%%%%%%%%%%%%%%%%%%%%%%%%%%%%%%%%
%%%%%%%%%%%%%%%%%%%%%%%%%%%%%%%%%%%%%%%%
% Table 4
\begin{table*}[tb]
\setlength{\tabcolsep}{5pt}
\renewcommand{\arraystretch}{1.05}
\centering
\small
\begin{tabular}{@{}|@{~}l|c|c|c|c|c|c|c|c|c|c|c|c|c|c|c|c|c|@{}}
\hline
\textbf{Model} & \textbf{Training} & \multicolumn{2}{c|}{\textbf{BC5CDR-Chemical}} & \multicolumn{2}{c|}{\textbf{BC5CDR-Disease}} & \multicolumn{2}{c|}{\textbf{BC4CHEMD}} & \multicolumn{2}{c|}{\textbf{NCBI-Disease}} \\ \hline

% & O + K &  (O) &  (K) &  (O) &  (K) &  (O) &  (K) &  (O) &  (K)  \\
% BERT & O + W &  (O) &  (W) &  (O) &  (W) &  (O) &  (W) &  (O) &  (W) \\
% & O + S &  (O) &  (S) &  (O) &  (S) &  (O) &  (S) &  (O) &  (S)  \\ \hline

 & O + K & .934 (O) & .888 (K) & .863 (O) & .755 (K) & .920 (O) & .874 (K) & .886 (O) & .820 (K)  \\
BioBERT & O + W & .931 (O) & .899 (W) & .865 (O) & .781 (W) & .922 (O) & .892 (W) & .872 (O) & .848 (W) \\ 
 & O + S & .933 (O) & .910 (S) & .840 (O) & .819 (S) & .919 (O) & .923 (S) & .874 (O) & .875 (S)  \\\hline
     
 & O + K & .898 (O) & .820 (K) & .844 (O) & .717 (K) & .819 (O) & .750 (K) & .789 (O) & .668 (K)  \\
BlueBERT  & O + W & .896 (O) & .656 (W) & .841 (O) & .759 (W) & .818 (O) & .785 (W) & .784 (O) & .729 (W) \\ 
 & O + S & .900 (O) & .890 (S) & .818 (O) & .814 (S) & .820 (O) & .788 (S) & .773 (O) & .804 (S)  \\\hline

% & O + K & .882 (O) & .777 (K) & .803 (O) & .596 (K) & .865 (O) & .740 (K) & .833 (O) & .739 (K)  \\
% Elmo & O + W & .882 (O) & .768 (W) & .797 (O) & .606 (W) & .869 (O) & .776 (W) & .845 (O) & .768 (W) \\
% & O + S & .875 (O) & .862 (S) & .791 (O) & .742 (S) & .866 (O) & .851 (S) & .852 (O) & .827 (S)  \\ \hline

& O + K & .923 (O) & .870 (K) & .833 (O) & .732 (K) & .912 (O) & .837 (K) & .864 (O) & .820 (K)  \\
BioELMo & O + W & .922 (O) & .825 (W) & .838 (O) & .654 (W) & .913 (O) & .820 (W) & .875 (O) & .777 (W) \\
& O + S & .919 (O) & .901 (S) & .826 (O) & .799 (S) & .912 (O) & .901 (S) & .871 (O) & .848 (S)  \\ \hline

& O + K & .910 (O) & .859 (K) & .823 (O) & .713 (K) & .898 (O) & .828 (K) & .860 (O) & .793 (K)  \\
ChemPatent ELMo & O + W & .907 (O) & .835 (W) & .813 (O) & .682 (W) & .899 (O) & .824 (W) & .863 (O) & .804 (W) \\
& O + S & .904 (O) & .895 (S) & .813 (O) & .757 (S) & .895 (O) & .874 (S) & .848 (O) & .819 (S)  \\ \hline

% & O + K &  (O) &  (K) &  (O) &  (K) &  (O) &  (K) &  (O) &  (K)  \\
% W2V & O + W &  (O) &  (W) &  (O) &  (W) &  (O) &  (W) &  (O) &  (W) \\
% & O + S &  (O) &  (S) &  (O) &  (S) &  (O) &  (S) &  (O) &  (S)  \\ \hline

 & O + K & .888 (O) & .467 (K) & .773 (O) & .303 (K) & .832 (O) & .486 (K) & .820 (O) & .543 (K)  \\
BioMedical W2V  & O + W & .873 (O) & .598 (W) & .796 (O) & .482 (W) & .836 (O) & .609 (W) & .819 (O) & .639 (W) \\ 
 & O + S & .867 (O) & .883 (S) & .781 (O) & .787 (S) & .837 (O) & .852 (S) & .836 (O) & .804 (S)  \\\hline
 
 & O + K & .867 (O) & .454 (K) & .768 (O) & .307 (K) & .817 (O) & .482 (K) & .822 (O) & .548 (K)  \\
ChemPatent W2V  & O + W & .785 (O) & .619 (W) & .765 (O) & .477 (W) &  .819 (O) & .626 (W) & .792 (O) & .663 (W) \\ 
 & O + S & .868 (O) & .864 (S) & .738 (O) & .779 (S) & .818 (O) & .835 (S) & .797 (O) & .801 (S)  \\\hline
\end{tabular}
\vspace{-3mm}
\caption{\label{table-4} Adversarial training results in terms of F1-score for each model and dataset. The training column shows the \textbf{O} set merged with \textbf{K}, \textbf{W}, or \textbf{S}. The test set is shown in parentheses for each scenario.
% The train set is shown as the original train set plus each of the adversarial train sets, keyboard, swap, and synonym.  
}
\vspace{-4mm}
\end{table*}
%%%%%%%%%%%%%%%%%%%%%%%%%%%%%%%%%%%%%%%%
%%%%%%%%%%%%%%%%%%%%%%%%%%%%%%%%%%%%%%%%

%%%%%%%%%%%%%%%%%%%%%%%%%%%%%%%%%%%%%%%%
\paragraph{Adversarial Evaluation Results} 
%%%%%%%%%%%%%%%%%%%%%%%%%%%%%%%%%%%%%%%%
Table~\ref{table-3} shows the evaluation results on the original (\textbf{O}) and adversarial test sets (\textbf{K}, \textbf{W}, and \textbf{S}). In general, the performance of models drops across all adversarial attacks. % As expected, the biomedical models performed better than their non-bio counterparts. 
For BERT-based models, we observe that \textbf{K} attacks decrease performance by on average 43.1\%, \textbf{W} by 34.3\% and \textbf{S} by 30.8\%. BioBERT has the smallest decrease in performance, 34.4\%, followed by BlueBERT, with a 37.9\% decrease. We hypothesize that BioBERT is more robust than BlueBERT since the former was trained on a larger and more varied corpus. Furthermore, when comparing the performance across all datasets, we see that \textbf{BC5CDR-Disease} is the most affected in all stress tests, with a 37.7\% performance drop, and the least affected is \textbf{BC5CDR-Chemical}, with 16.1\%.

The performance reduction of ELMo-based models is similar to those of BERT-based models. An exception is when subject to \textbf{W} and \textbf{S} noise, where 
they showed increased robustness with respect to BERT and W2V models (\textbf{W}: 55.3\% better, \textbf{S}: 6.9\% better). In almost all the tests, BioELMo performed better than ChemPatent ELMo, except under \textbf{W} noise, where ChemPatent ELMo performed consistently better, by 5.1\% on average. We hypothesize that these results are due to ELMo using a character-based input representation, which would allow handling of swap characters inside the words. 

W2V-based models were the most brittle but showed similar patterns to the previous models. Adversaries examples produced performance drops ranging from 53.8\% on \textbf{NCBI-Disease} to 74.1\% on \textbf{BC5CDR-Disease}. In the case of \textbf{S} adversaries, W2V-based showed performance drops ranging from 17.8\% on \textbf{BC5CDR-Chemical} to 55.3\% on \textbf{BC5CDR-Disease}.

Regarding the ``vanilla'' models, we see that they are all the worst in the original dataset (\textbf{O}) compared to their biomedical counterparts. In the same way, they are more fragile to adversary attacks in the biomedical scenario. In average, BERT has a decrease in performance of 39.6\%, ELMo of 34.4\% and W2V of 59.6\% across all datasets.

Even though the \textbf{BC5CDR} dataset covers both chemicals and diseases, the disease task is more affected by \textbf{S} adversaries. We believe this is due to the higher number of words affected by the attacks compared to
the other benchmarks (Table~\ref{table-2}). Another possible cause is the kind of synonyms used to replace the entities, which tend to be both superficially dissimilar and more extensive than their originals, e.g., \textit{arrhythmia} is replaced by \textit{heart conduction disorder}. By contrast, chemical synonyms often include terms derived from the original, e.g., \textit{morphine} is changed to \textit{morphine sulfate}.

%%%%%%%%%%%%%%%%%%%%%%%%%%%%%%%%%%%%%%%%
\paragraph{Training on Adversarial Examples} 
Additionally, we subjected the training sets to adversarial attacks, and evaluated the models both against the original test sets and their noisy counterparts.
When training with \textbf{K} noise, we observed performance decreases by 21.2\%, followed by \textbf{W}, 15.8\%, and \textbf{S} with a slight decline of 0.8\%, compared to 44.4\%, 46.3\% and 31.3\% respectively in the Adversarial Evaluation setting. Besides, and interestingly, training with \textbf{S} improves performance in some cases, by up to 5.5\% compared to the original \textbf{S} test set.
We hypothesize that this is because the introduced adversarial samples work as a data augmentation mechanism.
In terms of datasets, we see that \textbf{BC5CDR-Disease} is the most affected by adversaries, with an average 17.5\% drop, and the least affected is \textbf{NCBI-Disease}, with an average 9.7\% drop compared to the non-adversarial test set.
When comparing the three architectures we see that BERT is affected by 6.3\%, ELMo by 7.6\% and W2V by 24.0\% on average compared to the original test set. This result stands in line with findings on other NLP tasks, where BERT comes up first, followed by ELMo and W2V \cite{peng-etal-2019-transfer}.
This is because BERT uses recent methods and techniques like Transformer \cite{vaswani2017attention} and WordPiece tokenizer \cite{6289079} that allow it to learn better representations.
%Contextualized embeddings usually contain an input character encoding layer that makes them less sensitive to typos. 
% In general, we see that the BERT models are the ones that benefit the most. It could be because these models use the WordPiece tokenizer
% \cite{6289079} 
%which minimizes the number of words not seen during training.
% Word piece tokens cover almost all the words, even the words that do not occur in the dictionary. 
%When comparing the three architectures we see that BERT is affected by 6.3\%, ELMo by 7.6\% and W2V by 24.0\% on average.
%In terms of datasets, we see that \textbf{BC5CDR-Disease} is the most hurt by adversaries, with an average 17.5\% drop, and the least affected is \textbf{NCBI-Disease}, with an average 9.7\% drop compared to the non-adversarial dataset.

%%%%%%%%%%%%%%%%%%%%%%%%%%%%%%%%%%%%%%%%
\paragraph{BioBERT Error Analysis}
%%%%%%%%%%%%%%%%%%%%%%%%%%%%%%%%%%%%%%%%

%%%%%%%%%%%%%%%%%%%%%%%%%%%%%%%%%%%%%%%%
\begin{figure}
    \centering
\vspace{-2mm}        
    \subfigure[]{\includegraphics[width=0.37\textwidth]{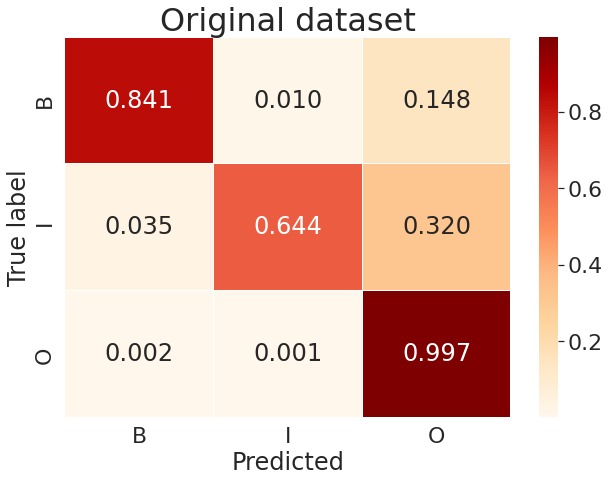}    \label{fig:1}} 
    \subfigure[]{\includegraphics[width=0.37\textwidth]{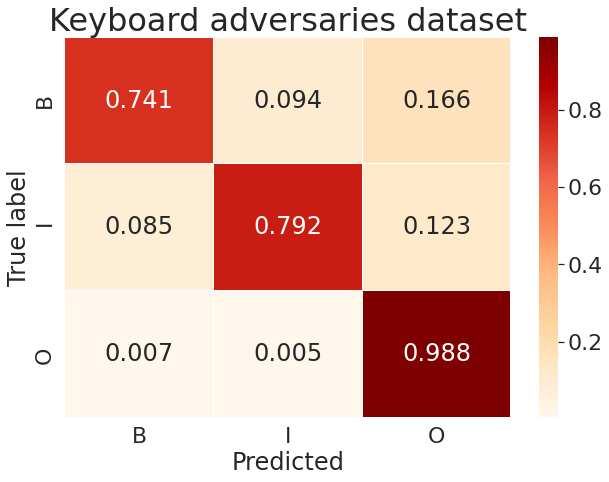}
    \label{fig:2}} 
    \subfigure[]{\includegraphics[width=0.37\textwidth]{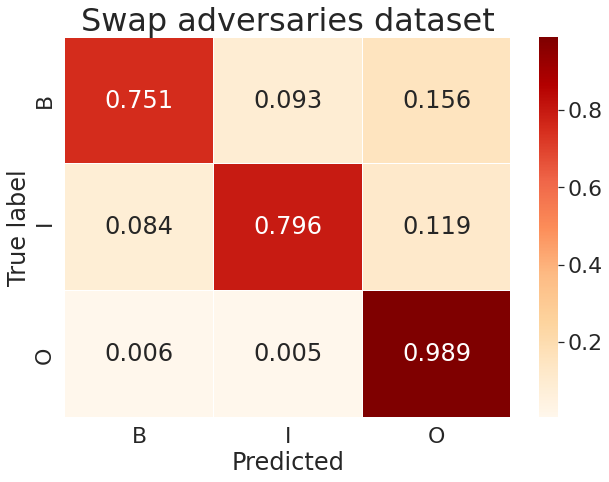}
    \label{fig:3}}
    \subfigure[]{\includegraphics[width=0.37\textwidth]{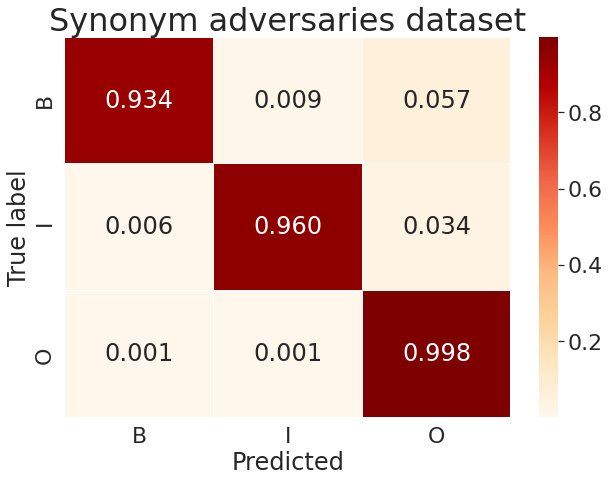}
    \label{fig:4}}
    \caption{Normalized confusion matrices for test results with (a) original (\textbf{O}), (b) keyboard (\textbf{K}), (c) swap  (\textbf{S}) and (d) synonym (\textbf{S}) BC5CDR-Disease and Chemical datasets on average.}
    \label{confusion_matrices}
\vspace{-4mm}    
\end{figure}
\vspace{-2mm}        
%%%%%%%%%%%%%%%%%%%%%%%%%%%%%%%%%%%%%%%%

This section seeks to understand how the most robust model -- BioBERT -- behaves under adversarial evaluation. To this end, we analyzed NER model confusions with respect to the original datasets, synonym (\textbf{S}), swap (\textbf{W}), and keyboard (\textbf{K}) perturbations on the BC5CDR chemical and disease dataset(s).

% \begin{figure}[H]
%     \centering
%     \includegraphics[width=\linewidth]{original_mat.png}
% \label{original_confusion_mat}
% \end{figure}

% \begin{figure}[tb]
%     \includegraphics[width=\linewidth]{keyboard_mat.png}
% \label{keyboard_confusion_mat}
% \end{figure}

% \begin{figure}[tb]
%     \includegraphics[width=\linewidth]{swap_mat.png}
% \label{swap_confusion_mat}
% \end{figure}

% \begin{figure}[tb]
%     \centering
%     \includegraphics[width=\linewidth]{synonym_mat.png}
% \label{synonym_confusion_mat}
% \end{figure}

In the original dataset (Figure \ref{fig:1}), we see that most of the errors come from confusing I and O labels (32\% of the cases). Under adversarial attacks, this type of error spreads to other IOB labels. For keyboard (\textbf{K}) errors (Figure \ref{fig:2}), the most frequent mistake is to confuse B with O, with 16.6\% of these cases. The same goes for swap (\textbf{W}) perturbations (Figure \ref{fig:3}), where this error is repeated 15\% of the time. When using synonyms (\textbf{S}) (Figure \ref{fig:4}), error rates become by contrast globally low compared to \textbf{K} and \textbf{W}. 
% We believe this happens because when perturbing synonyms, multi-token entities' meanings get compromised.
We believe that this happens because entities are converted into similar ones. For instance, ``stomach neoplasm'' gets transformed into ``stomach tumor''.
% viz., a precursor of cancer, but not cancer itself. 

Lastly, regardless of the adversaries, there are confusions with numbers and special character sequences that the model classifies as I (i.e., lie inside an entity span) but whose ground truth label is O (i.e., lie outside an entity span).

%%%%%%%%%%%%%%%%%%%%%%%%%%%%%%%%%%%%%%%%
\section{Conclusions} 
%%%%%%%%%%%%%%%%%%%%%%%%%%%%%%%%%%%%%%%%
In this work, we have investigated whether large scale biomedical word (W2V) and contextualized word embeddings (BERT and ELMo) are robust with respect to black-box adversarial attacks in the biomedical NER task. Our experimental results show different sensitivities of the models to misspellings and synonyms. 
Among the main findings, we show that BERT-based models are generally better prepared for adversarial attacks, but they are still fragile, leaving room for future improvement in the field. ELMo-based models show lower robustness in most cases but consistently outperformed BERT in some specific scenarios. W2V proves to be more brittle but shows similar patterns in terms of relative performance drops.
We also demonstrate that by training with adversaries, we can considerably decrease the drop in performance 
%of the models tested with noisy data 
and even improve the models' original performance when trained with synonyms, as they act as a form of regularization and augmentation of data. 
 
\section*{Acknowledgements}
We are grateful to the anonymous reviewers for their valuable feedback on earlier versions of this paper.
This work was partially funded by ANID - Millennium Science Initiative Program - Code ICN17\_002 and by ANID, FONDECYT grant 1191791, as well as supported by the TPU Research Cloud (TRC) program.



\begin{thebibliography}{37}
\expandafter\ifx\csname natexlab\endcsname\relax\def\natexlab#1{#1}\fi

\bibitem[{Agirre and Soroa(2007)}]{agirre2007semeval}
Eneko Agirre and Aitor Soroa. 2007.
\newblock Semeval-2007 task 02: Evaluating word sense induction and
  discrimination systems.
\newblock In \emph{Proceedings of the fourth international workshop on semantic
  evaluations (semeval-2007)}, pages 7--12.

\bibitem[{Akhtar and Mian(2018)}]{Akhtar_2018}
Naveed Akhtar and Ajmal Mian. 2018.
\newblock Threat of adversarial attacks on deep learning in computer vision: A
  survey.
\newblock \emph{IEEE Access}, 6:14410–14430.

\bibitem[{Araujo et~al.(2020)Araujo, Carvallo, and Parra}]{winlp}
Vladimir Araujo, Andr{\'e}s Carvallo, and Denis Parra. 2020.
\newblock \href
  {http://www.winlp.org/wp-content/uploads/2020/final_papers/34_Paper.pdf}
  {Adversarial evaluation of bert for biomedical named entity recognition}.
\newblock In \emph{Proceedings of the The Fourth Widening Natural Language
  Processing Workshop}, Seattle, USA. Association for Computational
  Linguistics.

\bibitem[{Aspillaga et~al.(2020)Aspillaga, Carvallo, and
  Araujo}]{aspillaga-etal-2020-stress}
Carlos Aspillaga, Andr{\'e}s Carvallo, and Vladimir Araujo. 2020.
\newblock \href {https://www.aclweb.org/anthology/2020.lrec-1.232} {Stress test
  evaluation of transformer-based models in natural language understanding
  tasks}.
\newblock In \emph{Proceedings of The 12th Language Resources and Evaluation
  Conference}, pages 1882--1894, Marseille, France. European Language Resources
  Association.

\bibitem[{Belinkov and Bisk(2018)}]{belinkov2018synthetic}
Yonatan Belinkov and Yonatan Bisk. 2018.
\newblock \href {https://openreview.net/forum?id=BJ8vJebC-} {Synthetic and
  natural noise both break neural machine translation}.
\newblock In \emph{International Conference on Learning Representations}.

\bibitem[{Bodenreider(2004)}]{Bodenreider2004}
O.~Bodenreider. 2004.
\newblock \href {https://doi.org/10.1093/nar/gkh061} {The unified medical
  language system ({UMLS}): integrating biomedical terminology}.
\newblock \emph{Nucleic Acids Research}, 32(90001):267D--270.

\bibitem[{Carvallo et~al.(2020)Carvallo, Parra, Lobel, and Soto}]{Carvallo2020}
Andres Carvallo, Denis Parra, Hans Lobel, and Alvaro Soto. 2020.
\newblock \href {https://doi.org/10.1007/s11192-020-03648-6} {Automatic
  document screening of medical literature using word and text embeddings in an
  active learning setting}.
\newblock \emph{Scientometrics}, 125(3):3047--3084.

\bibitem[{Devlin et~al.(2019)Devlin, Chang, Lee, and
  Toutanova}]{devlin-etal-2019-bert}
Jacob Devlin, Ming-Wei Chang, Kenton Lee, and Kristina Toutanova. 2019.
\newblock \href {https://doi.org/10.18653/v1/N19-1423} {{BERT}: Pre-training of
  deep bidirectional transformers for language understanding}.
\newblock In \emph{Proceedings of the 2019 Conference of the North {A}merican
  Chapter of the Association for Computational Linguistics: Human Language
  Technologies, Volume 1 (Long and Short Papers)}, pages 4171--4186,
  Minneapolis, Minnesota. Association for Computational Linguistics.

\bibitem[{Do{\u{g}}an et~al.(2014)Do{\u{g}}an, Leaman, and Lu}]{Doan2014}
Rezarta~Islamaj Do{\u{g}}an, Robert Leaman, and Zhiyong Lu. 2014.
\newblock \href {https://doi.org/10.1016/j.jbi.2013.12.006} {{NCBI} disease
  corpus: A resource for disease name recognition and concept normalization}.
\newblock \emph{Journal of Biomedical Informatics}, 47:1--10.

\bibitem[{Finlayson et~al.(2019)Finlayson, Bowers, Ito, Zittrain, Beam, and
  Kohane}]{Finlayson2019}
Samuel~G. Finlayson, John~D. Bowers, Joichi Ito, Jonathan~L. Zittrain,
  Andrew~L. Beam, and Isaac~S. Kohane. 2019.
\newblock Adversarial attacks on medical machine learning.
\newblock \emph{Science}, 363(6433):1287--1289.

\bibitem[{Goodfellow et~al.(2014)Goodfellow, Shlens, and
  Szegedy}]{goodfellow2014explaining}
Ian~J Goodfellow, Jonathon Shlens, and Christian Szegedy. 2014.
\newblock Explaining and harnessing adversarial examples.
\newblock \emph{arXiv preprint arXiv:1412.6572}.

\bibitem[{Hanahan and Weinberg(2000)}]{hoc2000}
Douglas Hanahan and Robert~A. Weinberg. 2000.
\newblock The hallmarks of cancer.
\newblock \emph{Cell 100(1)}, pages 57--70.

\bibitem[{Jean-Baptiste et~al.(2015)Jean-Baptiste, Alain, and Catherine}]{Lamy}
Lamy Jean-Baptiste, Venot Alain, and Duclos Catherine. 2015.
\newblock \href {https://doi.org/10.3233/978-1-61499-512-8-924} {Pymedtermino:
  an open-source generic api for advanced terminology services}.
\newblock \emph{Studies in Health Technology and Informatics}, 210(Digital
  Healthcare Empowering Europeans):924–928.

\bibitem[{Jia and Liang(2017)}]{jia-liang-2017-adversarial}
Robin Jia and Percy Liang. 2017.
\newblock \href {https://doi.org/10.18653/v1/D17-1215} {Adversarial examples
  for evaluating reading comprehension systems}.
\newblock In \emph{Proceedings of the 2017 Conference on Empirical Methods in
  Natural Language Processing}, pages 2021--2031, Copenhagen, Denmark.
  Association for Computational Linguistics.

\bibitem[{Jin et~al.(2019)Jin, Dhingra, Cohen, and Lu}]{jin2019probing}
Qiao Jin, Bhuwan Dhingra, William Cohen, and Xinghua Lu. 2019.
\newblock Probing biomedical embeddings from language models.
\newblock In \emph{Proceedings of the 3rd Workshop on Evaluating Vector Space
  Representations for NLP}, pages 82--89.

\bibitem[{Krallinger et~al.(2015)Krallinger, Rabal, Leitner, Vazquez, Salgado,
  Lu, Leaman, Lu, Ji, Lowe, Sayle, Batista-Navarro, Rak, Huber,
  Rockt\"{a}schel, Matos, Campos, Tang, Xu, Munkhdalai, Ryu, Ramanan, Nathan,
  {\v{Z}}itnik, Bajec, Weber, Irmer, Akhondi, Kors, Xu, An, Sikdar, Ekbal,
  Yoshioka, Dieb, Choi, Verspoor, Khabsa, Giles, Liu, Ravikumar, Lamurias,
  Couto, Dai, Tsai, Ata, Can, Usi{\'{e}}, Alves, Segura-Bedmar,
  Mart{\'{\i}}nez, Oyarzabal, and Valencia}]{Krallinger2015}
Martin Krallinger, Obdulia Rabal, Florian Leitner, Miguel Vazquez, David
  Salgado, Zhiyong Lu, Robert Leaman, Yanan Lu, Donghong Ji, Daniel~M Lowe,
  Roger~A Sayle, Riza~Theresa Batista-Navarro, Rafal Rak, Torsten Huber, Tim
  Rockt\"{a}schel, S{\'{e}}rgio Matos, David Campos, Buzhou Tang, Hua Xu,
  Tsendsuren Munkhdalai, Keun~Ho Ryu, SV~Ramanan, Senthil Nathan, Slavko
  {\v{Z}}itnik, Marko Bajec, Lutz Weber, Matthias Irmer, Saber~A Akhondi, Jan~A
  Kors, Shuo Xu, Xin An, Utpal~Kumar Sikdar, Asif Ekbal, Masaharu Yoshioka,
  Thaer~M Dieb, Miji Choi, Karin Verspoor, Madian Khabsa, C~Lee Giles, Hongfang
  Liu, Komandur~Elayavilli Ravikumar, Andre Lamurias, Francisco~M Couto,
  Hong-Jie Dai, Richard Tzong-Han Tsai, Caglar Ata, Tolga Can, Anabel
  Usi{\'{e}}, Rui Alves, Isabel Segura-Bedmar, Paloma Mart{\'{\i}}nez, Julen
  Oyarzabal, and Alfonso Valencia. 2015.
\newblock \href {https://doi.org/10.1186/1758-2946-7-s1-s2} {The {CHEMDNER}
  corpus of chemicals and drugs and its annotation principles}.
\newblock \emph{Journal of Cheminformatics}, 7(S1).

\bibitem[{Kringelum et~al.(2016)Kringelum, Kjaerulff, Brunak, Lund, Oprea, and
  Taboureau}]{chemprot2016}
J.~Kringelum, S.~K. Kjaerulff, S.~Brunak, O.~Lund, T.~I. Oprea, and
  O.~Taboureau. 2016.
\newblock Chemprot-3.0: a global chemical biology diseases mapping.
\newblock \emph{Database}.

\bibitem[{Lee et~al.(2019)Lee, Yoon, Kim, Kim, Kim, So, and
  Kang}]{10.1093/bioinformatics/btz682}
Jinhyuk Lee, Wonjin Yoon, Sungdong Kim, Donghyeon Kim, Sunkyu Kim, Chan~Ho So,
  and Jaewoo Kang. 2019.
\newblock \href {https://doi.org/10.1093/bioinformatics/btz682} {{BioBERT: a
  pre-trained biomedical language representation model for biomedical text
  mining}}.
\newblock \emph{Bioinformatics}, 36(4):1234--1240.

\bibitem[{Li et~al.(2016)Li, Sun, Johnson, Sciaky, Wei, Leaman, Davis,
  Mattingly, Wiegers, and Lu}]{Li2016}
Jiao Li, Yueping Sun, Robin~J. Johnson, Daniela Sciaky, Chih-Hsuan Wei, Robert
  Leaman, Allan~Peter Davis, Carolyn~J. Mattingly, Thomas~C. Wiegers, and
  Zhiyong Lu. 2016.
\newblock \href {https://doi.org/10.1093/database/baw068} {{BioCreative} v
  {CDR} task corpus: a resource for chemical disease relation extraction}.
\newblock \emph{Database}, 2016:baw068.

\bibitem[{Ma et~al.(2019)Ma, Niu, Gu, Wang, Zhao, Bailey, and
  Lu}]{ma2019understanding}
Xingjun Ma, Yuhao Niu, Lin Gu, Yisen Wang, Yitian Zhao, James Bailey, and Feng
  Lu. 2019.
\newblock \href {http://arxiv.org/abs/1907.10456} {Understanding adversarial
  attacks on deep learning based medical image analysis systems}.

\bibitem[{Mikolov et~al.(2013)Mikolov, Sutskever, Chen, Corrado, and
  Dean}]{NIPS2013_5021}
Tomas Mikolov, Ilya Sutskever, Kai Chen, Greg~S Corrado, and Jeff Dean. 2013.
\newblock \href
  {http://papers.nips.cc/paper/5021-distributed-representations-of-words-and-phrases-and-their-compositionality.pdf}
  {Distributed representations of words and phrases and their
  compositionality}.
\newblock In C.~J.~C. Burges, L.~Bottou, M.~Welling, Z.~Ghahramani, and K.~Q.
  Weinberger, editors, \emph{Advances in Neural Information Processing Systems
  26}, pages 3111--3119. Curran Associates, Inc.

\bibitem[{Naik et~al.(2018)Naik, Ravichander, Sadeh, Rose, and
  Neubig}]{naik-etal-2018-stress}
Aakanksha Naik, Abhilasha Ravichander, Norman Sadeh, Carolyn Rose, and Graham
  Neubig. 2018.
\newblock \href {https://www.aclweb.org/anthology/C18-1198} {Stress test
  evaluation for natural language inference}.
\newblock In \emph{Proceedings of the 27th International Conference on
  Computational Linguistics}, pages 2340--2353, Santa Fe, New Mexico, USA.
  Association for Computational Linguistics.

\bibitem[{Neumann et~al.(2019)Neumann, King, Beltagy, and
  Ammar}]{Neumann2019ScispaCyFA}
Mark Neumann, Daniel King, Iz~Beltagy, and Waleed Ammar. 2019.
\newblock \href {http://arxiv.org/abs/arXiv:1902.07669} {Scispacy: Fast and
  robust models for biomedical natural language processing}.
\newblock In \emph{SciSpacy:Fast and Robust Models for Biomedical Natural
  Language Processing}.

\bibitem[{Paschali et~al.(2018)Paschali, Conjeti, Navarro, and
  Navab}]{Paschali2018}
Magdalini Paschali, Sailesh Conjeti, Fernando Navarro, and Nassir Navab. 2018.
\newblock \href {https://doi.org/10.1007/978-3-030-00928-1_56}
  {Generalizability vs. robustness: Investigating medical imaging networks
  using adversarial examples}.
\newblock In \emph{Medical Image Computing and Computer Assisted Intervention
  {\textendash} {MICCAI} 2018}, pages 493--501. Springer International
  Publishing.

\bibitem[{Peng et~al.(2019)Peng, Yan, and Lu}]{peng-etal-2019-transfer}
Yifan Peng, Shankai Yan, and Zhiyong Lu. 2019.
\newblock \href {https://doi.org/10.18653/v1/W19-5006} {Transfer learning in
  biomedical natural language processing: An evaluation of {BERT} and {ELM}o on
  ten benchmarking datasets}.
\newblock In \emph{Proceedings of the 18th BioNLP Workshop and Shared Task},
  pages 58--65, Florence, Italy. Association for Computational Linguistics.

\bibitem[{Peters et~al.(2018)Peters, Neumann, Iyyer, Gardner, Clark, Lee, and
  Zettlemoyer}]{peters2018deep}
Matthew Peters, Mark Neumann, Mohit Iyyer, Matt Gardner, Christopher Clark,
  Kenton Lee, and Luke Zettlemoyer. 2018.
\newblock \href {https://doi.org/10.18653/v1/N18-1202} {Deep contextualized
  word representations}.
\newblock In \emph{Proceedings of the 2018 Conference of the North {A}merican
  Chapter of the Association for Computational Linguistics: Human Language
  Technologies, Volume 1 (Long Papers)}, pages 2227--2237, New Orleans,
  Louisiana. Association for Computational Linguistics.

\bibitem[{Pletscher-Frankild et~al.(2015)Pletscher-Frankild, Pallej{\`a},
  Tsafou, Binder, and Jensen}]{pletscher2015diseases}
Sune Pletscher-Frankild, Albert Pallej{\`a}, Kalliopi Tsafou, Janos~X Binder,
  and Lars~Juhl Jensen. 2015.
\newblock Diseases: Text mining and data integration of disease--gene
  associations.
\newblock \emph{Methods}, 74:83--89.

\bibitem[{Pyysalo et~al.(2013)Pyysalo, Ginter, Moen, Salakoski, and
  Ananiadou}]{Pyysalo:2013b}
S.~Pyysalo, F.~Ginter, H.~Moen, T.~Salakoski, and S.~Ananiadou. 2013.
\newblock \href {http://lbm2013.biopathway.org/lbm2013proceedings.pdf}
  {Distributional semantics resources for biomedical text processing}.
\newblock In \emph{Proceedings of LBM 2013}, pages 39--44.

\bibitem[{Ramshaw and Marcus(1999)}]{ramshaw1999text}
Lance~A Ramshaw and Mitchell~P Marcus. 1999.
\newblock Text chunking using transformation-based learning.
\newblock In \emph{Natural language processing using very large corpora}, pages
  157--176. Springer.

\bibitem[{{Schuster} and {Nakajima}(2012)}]{6289079}
M.~{Schuster} and K.~{Nakajima}. 2012.
\newblock Japanese and korean voice search.
\newblock In \emph{2012 IEEE International Conference on Acoustics, Speech and
  Signal Processing (ICASSP)}, pages 5149--5152.

\bibitem[{So{\u{g}}anc{\i}o{\u{g}}lu et~al.(2017)So{\u{g}}anc{\i}o{\u{g}}lu,
  {\"O}zt{\"u}rk, and {\"O}zg{\"u}r}]{souganciouglu2017biosses}
Gizem So{\u{g}}anc{\i}o{\u{g}}lu, Hakime {\"O}zt{\"u}rk, and Arzucan
  {\"O}zg{\"u}r. 2017.
\newblock Biosses: a semantic sentence similarity estimation system for the
  biomedical domain.
\newblock \emph{Bioinformatics}, 33(14):i49--i58.

\bibitem[{Sun et~al.(2018)Sun, Tang, Yi, Wang, and
  Zhou}]{10.1145/3219819.3219909}
Mengying Sun, Fengyi Tang, Jinfeng Yi, Fei Wang, and Jiayu Zhou. 2018.
\newblock \href {https://doi.org/10.1145/3219819.3219909} {Identify susceptible
  locations in medical records via adversarial attacks on deep predictive
  models}.
\newblock In \emph{Proceedings of the 24th ACM SIGKDD International Conference
  on Knowledge Discovery and Data Mining}, KDD ’18, page 793–801, New York,
  NY, USA. Association for Computing Machinery.

\bibitem[{Szegedy et~al.(2014)Szegedy, Zaremba, Sutskever, Bruna, Erhan,
  Goodfellow, and Fergus}]{42503}
Christian Szegedy, Wojciech Zaremba, Ilya Sutskever, Joan Bruna, Dumitru Erhan,
  Ian Goodfellow, and Rob Fergus. 2014.
\newblock \href {http://arxiv.org/abs/1312.6199} {Intriguing properties of
  neural networks}.
\newblock In \emph{International Conference on Learning Representations}.

\bibitem[{Tari et~al.(2010)Tari, Anwar, Liang, Cai, and
  Baral}]{tari2010discovering}
Luis Tari, Saadat Anwar, Shanshan Liang, James Cai, and Chitta Baral. 2010.
\newblock Discovering drug--drug interactions: a text-mining and reasoning
  approach based on properties of drug metabolism.
\newblock \emph{Bioinformatics}, 26(18):i547--i553.

\bibitem[{Vaswani et~al.(2017)Vaswani, Shazeer, Parmar, Uszkoreit, Jones,
  Gomez, Kaiser, and Polosukhin}]{vaswani2017attention}
Ashish Vaswani, Noam Shazeer, Niki Parmar, Jakob Uszkoreit, Llion Jones,
  Aidan~N Gomez, {\L}ukasz Kaiser, and Illia Polosukhin. 2017.
\newblock Attention is all you need.
\newblock In \emph{Advances in neural information processing systems}, pages
  5998--6008.

\bibitem[{Zhai et~al.(2019)Zhai, Nguyen, Akhondi, Thorne, Druckenbrodt, Cohn,
  Gregory, and Verspoor}]{zhai2019improving}
Zenan Zhai, Dat~Quoc Nguyen, Saber Akhondi, Camilo Thorne, Christian
  Druckenbrodt, Trevor Cohn, Michelle Gregory, and Karin Verspoor. 2019.
\newblock \href {https://doi.org/10.18653/v1/W19-5035} {Improving chemical
  named entity recognition in patents with contextualized word embeddings}.
\newblock In \emph{Proceedings of the 18th BioNLP Workshop and Shared Task},
  pages 328--338, Florence, Italy. Association for Computational Linguistics.

\bibitem[{Zhang et~al.(2019)Zhang, Sheng, Alhazmi, and
  Li}]{zhang2019adversarial}
Wei~Emma Zhang, Quan~Z. Sheng, Ahoud Alhazmi, and Chenliang Li. 2019.
\newblock \href {http://arxiv.org/abs/1901.06796} {Adversarial attacks on deep
  learning models in natural language processing: A survey}.

\end{thebibliography}
\end{document}